\documentclass[pdflatex,sn-mathphys-num]{sn-jnl}
\usepackage{graphicx}%
\usepackage{multirow}%
\usepackage{amsmath,amssymb,amsfonts}%
\usepackage{amsthm}%
\usepackage{mathrsfs}%
\usepackage[title]{appendix}%
\usepackage{xcolor}%
\usepackage{textcomp}%
\usepackage{manyfoot}%
\usepackage{booktabs}%
\usepackage{algorithm}%
\usepackage{algorithmicx}%
\usepackage{algpseudocode}%
\usepackage{listings}%
\usepackage{xcolor} 

\theoremstyle{thmstyleone}%
\theoremstyle{thmstyletwo}%

\theoremstyle{thmstylethree}%

\begin{document}

\title[Article Title]{RW-NSGCN: A Robust Approach to Structural Attacks via Negative Sampling}


\author[1]{\fnm{Shuqi} \sur{He}}
\author[2]{\fnm{Jun} \sur{Zhuang}}
\author[1]{\fnm{Ding} \sur{Wang}}
\author*[1]{\fnm{Jun} \sur{Song}}

\affil[1]{\orgdiv{Computer Science and technology}, \orgname{China University of Geosciences (Wuhan)}, \orgaddress{\city{Wuhan}, \postcode{45000}, \state{Hubei}, \country{China}}}

\affil[2]{Independent Researcher}


\abstract{Node classification using Graph Neural Networks (GNNs) has been widely applied in various practical scenarios, such as predicting user interests and detecting communities in social networks. However, recent studies have shown that graph-structured networks often contain potential noise and attacks, in the form of topological perturbations and weight disturbances, which can lead to decreased classification performance in GNNs. To improve the robustness of the model, we propose a novel method:  Random Walk Negative Sampling Graph Convolutional Network (RW-NSGCN). Specifically, RW-NSGCN integrates the Random Walk with Restart (RWR) and PageRank (PGR) algorithms for negative sampling and employs a Determinantal Point Process (DPP)-based GCN for convolution operations. RWR leverages both global and local information to manage noise and local variations, while PGR assesses node importance to stabilize the topological structure. The DPP-based GCN ensures diversity among negative samples and aggregates their features to produce robust node embeddings, thereby improving classification performance. Experimental results demonstrate that the RW-NSGCN model effectively addresses network topology attacks and weight instability, increasing the accuracy of anomaly detection and overall stability. In terms of classification accuracy, RW-NSGCN significantly outperforms existing methods, showing greater resilience across various scenarios and effectively mitigating the impact of such vulnerabilities.}




\maketitle

\section{Introduction}\label{sec1}
With advancements in Graph Neural Network (GNN) technology, GNNs are now extensively applied in the classification tasks of graph-structured networks~\cite{wu2020comprehensive, he2022explainer, zhuang2022deperturbation}. These networks effectively capture the features of nodes and edges within a graph, making them crucial tools for identifying complex patterns and relationships.
For instance, GNNs are widely used in the financial sectors for identifying the transaction connections in the networks~\cite{zhou2020graph, motie2023financial, yu2024credit}, which are commonly seen in our daily lives, such as E-commerce platforms~\cite{wu2023enhanced, xu2024f}, E-health systems~\cite{huang2024research, yu2024enhancing}, and real estate markets~\cite{yan2022influencing}. Despite the powerful capabilities of Graph Neural Networks (GNNs) in classification tasks, challenges remain due to the inherent topological Vulnerability~\cite{hussain2021structack, zou2021tdgia, zhuang2022defending} and weight instability~\cite{wu2023adversarial, wang2022pruning} of graph-structured networks.

\begin{itemize}
\item Topological Vulnerability refers to the significant impact on model output when there are minor changes in the node connections (i.e., the topology) of graph-structured data~\cite{shen2012discovery, trappolini2023sparse, zhuang2023robust}. For example, GNNs update node representations by aggregating information from neighboring nodes, where changes in node features may be propagated and amplified. Furthermore, in complex networks, multi-hop information propagation between nodes may lead to information loss or miscommunication due to slight topological changes.
\item Weight Sensitivity indicates that GNNs are highly responsive to variations in weight initialization and optimization processes~\cite{sharma2024investigating, xue2021cap, wu2023adversarial}. For instance, noise and outliers in the input data can impact weight updates, leading to unstable model performance. If the training data contains bias or noise, the model may overfit these undesirable patterns, resulting in increased weight sensitivity.
\end{itemize}

Existing methods for analyzing graph-structured networks, such as Graph Convolutional Networks (GCN)~\cite{Kipf_Welling_2016}, Graph Sampling and Aggregation (GraphSAGE)~\cite{Hamilton_Ying_Leskovec_2017}, and Graph Attention Networks (GATv2)~\cite{Brody_Alon_Yahav_2021}, primarily focus on aggregating data from neighboring nodes for risk assessment and pattern recognition. Although these methods are somewhat effective, they often overlook information from non-adjacent nodes, which can indirectly impact the network's overall flow and weight distribution. This localized perspective limits their ability to respond to topological changes and fluctuations in weight distribution. Therefore, understanding the intricate relationships between non-neighbor nodes is crucial for developing more accurate classification models. However, current research still faces limitations in addressing this issue. Some studies, like those on SDGCN~\cite{duan2023graph}, have introduced negative sampling mechanisms, but their random negative sampling methods fail to fully leverage the importance of nodes and global structural information. Consequently, these methods have limited capacity to capture complex relationships between non-adjacent nodes, hindering accurate assessment of nodes' global influence within the network. This, in turn, affects node representation and the overall stability of the model.

In response to the limitations of current graph structure network analysis methods, we propose a new approach called Random Walk Neural Sampling Graph Convolutional Network (RW-NSGCN) to improve the resilience of graph structure networks under attack. RW-NSGCN introduces a negative sampling mechanism based on the random walk algorithm to effectively utilize information from non-adjacent nodes~\cite{nikolentzos2020random,Jin2022RAWGNNRW}. By integrating the Random Walk with Restart algorithm and PageRank~\cite{bojchevski2020scaling,he2022gnnrank}-based negative sampling, this model combines global and local information, reduces noise interference, reassesses node importance, and identifies key nodes, thereby improving network stability.

To further improve robustness, RW-NSGCN aggregates information from both positive and negative samples through graph convolutional networks. It identifies significant relationships between nodes lacking direct connections using a shortest path-based method, revealing complex network links and facilitating the analysis of potential influences and information diffusion among nodes. This approach overcomes the challenges of analyzing non-adjacent node relationships in complex networks. Additionally, the model improves the diversity of negative samples through a Determinantal Point Process (DPP) based on node association strength~\cite{mariet2019learning,poulson2020high}, ensuring representativeness and coverage.

The experimental results show that the RW-NSGCN model not only outperforms other models in accuracy on various datasets but also shows outstanding robustness in experiments involving artificially induced topological and weight perturbations. Through ablation studies, this research confirms the effectiveness of random walk and PageRank techniques and conducts a thorough parameter analysis. Overall, RW-NSGCN shows significant advantages in handling complex graph data.

Our contributions are highlighted in the following three areas:

\begin{itemize}
    \item This paper introduces the RW-NSGCN model, which combines GCN with algorithms based on random walks. This integration improves network connectivity and robustness, addressing the topological fragility and weight instability inherent in graph-structured networks.
    \item We use DPP based on the strength of node association to ensure negative sample diversity and enhance feature robustness, and utilize the shortest path to preserve critical topological information and improve model robustness.
    \item The experimental results indicate that RW-NSGCN outperforms the current state-of-the-art models in classification accuracy across multiple datasets such as Cora. It also demonstrates strong robustness in scenarios with topological and weight perturbations.
\end{itemize}

\section{Related work}\label{sec2}
In recent years, employing graph neural networks on classification tasks has emerged as a prominent research focus~\cite{jiang2021tripoline, yin2022glign, zhuang2022does, liu2022joint, liu2021neural, mao2022trace, mo2024cross, mo2024pi, he2022earth, lu2024cats, zhuang2024robust}. It has been shown that graph neural networks (GNNs) are susceptible to topology changes and edge weight perturbations when processing graph data~\cite{lyu2023attention, mo2022trafficflowgan, he2024enhancing, zhuang2022robust}. Conventional methods are no longer applicable in these cases~\cite{skhiri2012large}. For instance, graph convolution models like GCN~\cite{Kipf_Welling_2016,gan2022multigraph} and GraphSAGE~\cite{Hamilton_Ying_Leskovec_2017, Chen2021Adversarial} capture information from adjacent nodes through graph convolution and reduce computational costs through sampling and aggregation, scaling up the representation learning capacity on large-scale dynamic environments~\cite{sun2023manifold}. Attention-based models, such as GATv2~\cite{Brody_Alon_Yahav_2021, Shi2021Boosting-GNN} and MAD~\cite{Chen_Lin_Li_Li_Zhou_Sun_2020}, dynamically adjust node weights to identify critical nodes and address graph-structured networks with dynamic edge weights. However, these models rely solely on the information transmission from neighboring nodes, overlooking information that impacts the entire network via indirect paths. This oversight limits the model's ability to effectively manage topological and weight perturbations. This is because perturbations often affect the global structure and overall edge weights, altering indirect paths that connect distant nodes. Consequently, the models become less robust to these changes and may fail to accurately capture critical long-range dependencies, leading to a decrease in classification performance.

Recent research highlights that incorporating a negative sampling mechanism into neural network models can significantly enhance their performance~\cite{yang2020understanding}, as noted by Duan et al.~\cite{duan2021negative,duan2022learning,duan2023graph}. NegGCN (MCGCN) employs Markov Chain Monte Carlo (MCMC)-based negative sampling, while D2GCN and SDGCN utilize Determinantal Point Processes (DPP). By introducing negative samples, these techniques more effectively capture the diverse node features present in graph-structured networks, thereby improving classification accuracy. Furthermore, negative sampling mitigates the over-smoothing problem in node representations within multi-layer models~\cite{elinas2022addressing,shen2024resisting}, enhancing the robustness of these models in graph-structured networks. However, the implementation of random negative sampling during the selection process~\cite{yoon2021performance,wei2022evaluating,gao2024towards} presents significant limitations. Random sampling lacks explicit guidance and a global perspective tailored to graph topology, leading to insufficient sample diversity, incomplete coverage, local bias, and ineffective identification of key nodes, which results in the oversight of important nodes~\cite{ying2019gnnexplainer,munikoti2022scalable,shan2024kpi}. This deficiency hinders the model's ability to capture essential nodes and edge features, reducing its robustness to structural and weight perturbations and ultimately degrading the overall performance of graph neural networks. To address these challenges, more sophisticated and guided sampling methods are required to ensure comprehensive coverage of critical features and accurate identification of significant nodes.

To address the shortcomings of existing models, we propose several improvements. First, to tackle the problem of neglecting non-neighboring node information, we enhance the Graph Convolutional Network (GCN) information propagation mechanism by incorporating non-neighbor nodes into the information aggregation process, effectively capturing network-influencing information through indirect paths~\cite{liu2021enhancing,yang2021attributes}. Second, we address the limitations of negative sampling by integrating Random Walk with Restart (RWR) and PageRank (PGR) algorithms to select negative samples. The RWR algorithm considers both global and local information during graph traversal, smoothing noise effects, while the PGR algorithm evaluates the global importance of each node by calculating its PageRank value. By combining RWR and PGR algorithms, our approach integrates global information and considers the importance of critical nodes and edges, significantly enhancing the model's robustness and generalization capability in graph-structured network environments.

\section{Preliminaries}\label{sec3}
In GCNs, node representations are updated through graph convolution operations, which combine the features of the nodes themselves and their neighboring nodes, capturing the information inherent in the graph structure~\cite{wu2020comprehensive}. The process of updating node representations can be expressed by the following formula:

\begin{equation}
x^{(l+1)} = \text{ReLU}\left( \mathbf{W}^{(l)} \cdot \text{normalize}(x^{(l)}, \mathbf{D}, \mathbf{A}) \right),
\end{equation}
where $ x^{(l)} $ is node representations at layer $ l $, and $ x^{(0)} = \mathbf{X} $ represents initial node features~\cite{kipf2016semi}. A common operation for normalizing the node representation $ x^{(l)} $ is:

\begin{equation}
\text{normalize}(x^{(l)}, \mathbf{D}, \mathbf{A}) = \mathbf{D}^{-1/2} \mathbf{A} \mathbf{D}^{-1/2} x^{(l)} 
\end{equation}

Here, $\mathbf{D}$ denotes the Degree Matrix, and $\mathbf{A}$ is the Adjacency Matrix. The matrix $\mathbf{W}^{(l)}$ is the weight matrix from layer $ l $ to layer $ l+1 $, representing a linear transformation, while $\text{ReLU}$ is the Rectified Linear Unit activation function, used to introduce non-linearity.

\section{menthod}\label{sec4}
This research focuses on the security challenges in graph-structured networks, particularly the topological vulnerabilities and weight instability when GNNs perform classification tasks on such networks. These issues can disrupt normal network behavior and structure, thereby reducing the model's classification ability~\cite{zang2023guap}. Thus, it is crucial to develop models capable of addressing the topological vulnerabilities and weight instability present in networks.

\subsection{Definition}
We define the graph-structured network as a graph $ \mathbf{G} = (\mathbf{V, E}) $, where $ V $ represents the set of nodes, $ V = \{v_1, v_2, \cdots, v_N\} $, with $ N $ being the number of nodes, and $ E $ is the set of edges representing the connections between nodes. An edge $ e \in E $ indicates a relationship between nodes.

\subsection{Overview}
In order to effectively solve the node distance measurement problem as well as the non-neighbor node selection problem in complex networks, we have developed a negative sample selection model based on the calculation of the combined score of the shortest path and random wandering. The model generates a set of non-neighboring nodes for varying path lengths by computing the shortest path lengths between nodes. Then, it selects candidate nodes for information aggregation from the non-neighbor nodes set based on RWR (Random Walk with Restart) and PGR (PageRank) scores. Finally, by integrating the DPP (Determinantal Point Process) kernel matrix with GCN (Graph Convolutional Network), we ensure the diversity and representativeness of the sampled nodes.

\subsection{Selection of Negative Samples}
We propose a method to identify and protect important nodes by calculating the shortest path and combining scores.

First, by calculating the shortest path length $d(v_i, v_j)$ between node pairs, we can determine the relationship chain length between two nodes. Then, we generate a set of reachable nodes for each node $v_i$ at different path lengths $l$:

\begin{equation}
\mathcal{N}_l(v_i) = \{ v_j \mid d(v_i, v_j) = l \}
\end{equation}

This step helps identify the set of nodes that have specific path relationships with node $v_i$, understand path diffusion, and analyze the node's position within the local structure and its relation to other nodes. Through these sets, we can further analyze the influence of nodes under different paths, which helps identify key nodes. The steps are shown in Algorithm~\ref{alg:short}.

\begin{algorithm}
\caption{Diverse Negative Sampling Based on Modified Shortest Paths}
\label{alg:short}
\begin{algorithmic}[1]
\Require A graph $G$, maximum sampling length $L_{max}$
\Ensure $S(i)$ and $N(i)$ for all $i \in G$
\For{each node $i$ in $G$}
    \State $S(i) \leftarrow \{\}$
    \State Compute the shortest path lengths from $i$ to all reachable nodes $V_r$
    \State $N(i) \leftarrow \{\}$
    \For{$len = 1$ to $L_{max}$}
        \State $R(len) \leftarrow$ Collect nodes with path length $len$ $N_{len}$
        \If{$len = L_{max}$ or $len = L_{max} - 1$}
            \State $S(i) \leftarrow S(i) \cup R(len)$
        \Else
            \State $S(i) \leftarrow S(i) \cup \{\text{Nei}(j) | j \in R(len)\}$
        \EndIf
        \State $N(i) \leftarrow N(i) \cup R(len)$
    \EndFor
\EndFor
\State \Return $S(i)$ and $N(i)$ for all $i \in G$
\end{algorithmic}
\end{algorithm}
By employing scoring vectors \( s_{ij} \) derived from the combination of Random Walk with Restart (RWR) and Personalized PageRank (PGR), our analysis capability is significantly enhanced. RWR, through multiple restarts and exploration of multiple paths, provides robust local information, aiding in identifying relationships between nodes and their neighbors. Meanwhile, the PGR scores reflect the global significance of nodes, enabling the identification of key nodes at any stage of the graph structure, ensuring no crucial nodes are overlooked.

\begin{equation}
    s_{ij} = \beta \cdot (1 - \alpha) \left[ (\mathbf{I} - \alpha \mathbf{P})^{-1} \mathbf{e}_i \right]_j + (1 - \beta) \left[ \alpha \mathbf{P} (\alpha \mathbf{P} \mathbf{p} + (1 - \alpha) \mathbf{e}) + (1 - \alpha) \mathbf{e} \right]_j \quad \forall v_j \in \mathcal{N}_l(v_i),
\end{equation}
where \(\mathbf{P} = \mathbf{D}^{-1} \mathbf{A}\) is the transition matrix, \(\alpha\) is the restart probability or damping factor, and \(\beta\) is the weighting factor for the RWR and PGR scores.

Based on these composite scoring vectors, we select the highest-scoring nodes from node sets of varying path lengths as candidate nodes:

\begin{equation}
   \text{candidates}(v_i) = \bigcup_{l=1}^{L} \text{top}(\{ s_{ij} \mid v_j \in \mathcal{N}_l(v_i) \})
\end{equation}

By integrating critical node information from non-neighbor nodes at different distances, nodes can capture multi-scale features and acquire diverse information from local to global perspectives, thereby more comprehensively representing their characteristics. Additionally, incorporating information from nodes at varying distances increases information diversity, avoiding the limitations of relying solely on local neighborhood information, thereby improving the model's robustness and generalization capability. When information from more distant nodes is included, the model can better understand the global structure, capture global patterns, and handle long-range dependencies, especially providing valuable context information when local information is insufficient. In summary, this approach effectively addresses the topological vulnerabilities and weight instability in graph-structured networks, enhancing overall stability and robustness.

\subsection{GCN Based on Label Propagation with DPP Sampling}
By integrating Determinantal Point Processes (DPP) with Graph Convolutional Networks (GCN), we effectively capture the feature associations between nodes and communities while ensuring node diversity, thus improving defense capabilities.

First, the model uses a DPP kernel matrix to measure the association strength between node pairs, defined as follows:
\begin{equation}
\mathbf{L} = \mathbf{Q} \cdot (\mathbf{S}_{com} \cdot \mathbf{S}_{com}^T) \cdot \exp(\mathbf{S}_{node} - 1) \cdot \mathbf{Q}^T
\end{equation}

In this context, $\mathbf{L}$ represents the DPP kernel matrix, where each element $(\mathbf{L})_{ij}$ signifies the association strength between nodes $v_i$ and $v_j$, both selected from $\text{candidates}(v)$. The $\mathbf{S}$ matrix consists of three types of similarity matrices:

\begin{equation}
\label{eqa:1}
(\mathbf{S}_{node})_{ij} = \cos(\mathbf{x}_i, \mathbf{x}_j) = \frac{\mathbf{x}_i \cdot \mathbf{x}_j}{\|\mathbf{x}_i\| \|\mathbf{x}_j\|}
\end{equation}

\begin{equation}
\label{eqa:2}
(\mathbf{S}_{com})_{ij} = \cos(\mathbf{f}_i, \mathbf{f}_j) = \frac{\mathbf{f}_i \cdot \mathbf{f}_j}{\|\mathbf{f}_i\| \|\mathbf{f}_j\|}
\end{equation}

\begin{equation}
\label{eqa:3}
\mathbf{Q}_{ij} = \cos(\mathbf{c}_i, \mathbf{f}_j) = \frac{\mathbf{c}_i \cdot \mathbf{f}_j}{\|\mathbf{c}_i\| \|\mathbf{f}_j\|}
\end{equation}

Equation~\ref{eqa:1} quantifies the similarity of nodes within the feature space, while Equation~\ref{eqa:2} measures the similarity between community features. These community features are aggregated from the attributes of all nodes within the community using a label propagation algorithm, facilitating a comprehensive understanding of relationships and resemblances between different communities. Equation~\ref{eqa:3} represents the similarity between the central node and the community features. These similarity matrices enable a thorough assessment of node association strength, capturing node similarity and correlation across multiple dimensions. The above steps are designed to effectively capture the feature relationships between nodes and communities, acquiring rich information between them.

After constructing the DPP kernel matrix $\mathbf{L}$, sampling is performed to ensure the selected node samples exhibit sufficient representativeness and diversity, avoiding bias and over-concentration in node selection.

\begin{equation}
\text{DPP.sample}(L, k)
\end{equation}

The obtained negative samples of DPP are mainly used in the GCN messaging phase:

\begin{equation}
x^{(l+1)}_{\text{pos}} = \text{ReLU}\left( \mathbf{W}^{(l)} \cdot \text{normalize}(x^{(l)}, \mathbf{D}, \mathbf{A}) \right)
\end{equation}

\begin{equation}
x^{(l+1)}_{\text{neg}} = \text{ReLU}\left( \mathbf{W}^{(l)}_{\text{DPP}} \cdot \text{normalize}(x^{(l)}, \mathbf{D}_{\text{DPP}}, \mathbf{A}_{\text{DPP}}) \right)
\end{equation}

\begin{equation}
x^{(l+1)} = x^{(l+1)}_{\text{pos}} - \lambda \cdot x^{(l+1)}_{\text{neg}}
\end{equation}

First, we use standard GCN convolution to propagate the features of positive samples, capturing the characteristics of normal behavior. Next, we introduce the DPP sampling mechanism to select negative sample nodes and propagate their features. Finally, node representations are updated by subtracting the features of negative samples (multiplied by a balance coefficient $\lambda$) from those of positive samples. This approach ensures that the selected node subset is discrete in the feature space, allowing the selection of nodes that are not too similar to each other, thus avoiding redundant information selection, effectively preventing overfitting, and enhancing sensitivity to anomalous behaviors in graph structures.

\section{experiment}\label{sec5}
This study explores the topological vulnerabilities and weight instability present in Graph Neural Networks (GNN) during classification tasks within graph-structured networks. To address these challenges, we propose a negative sampling method that combines random walk, PageRank, and Determinantal Point Processes (DPP) sampling to improve the robustness of graph neural networks. Experiments conducted on the Cora, CiteSeer, and PubMed datasets demonstrate the significant advantages of this new method in accuracy and node classification. We also performed comparative experiments to further evaluate the model's robustness in handling significant topological perturbations and weight interference. Additionally, we conducted ablation studies and sensitivity analyses to further explore the impact of different parameter settings on model performance. The results indicate a significant improvement in accuracy when handling complex graph data.

\subsection{Datasets}
Cora~\cite{sen2008collective}, CiteSeer, and PubMed~\cite{wheeler2007database} are three renowned citation network datasets widely used in machine learning and graph neural network research. Cora contains citation relationships between machine learning papers, CiteSeer covers computer science literature, and PubMed focuses on biomedical articles. These datasets serve as standard benchmarks for graph node classification and link prediction tasks. We tested the model's baseline performance on the Cora and CiteSeer datasets and extracted subgraphs from PubMed to evaluate weight perturbations and critical link attacks.
\begin{table}[ht]
    \centering
    \caption{Statistics of Cora, PubMed, and Citeseer Datasets}
    \begin{tabular}{lcccc}
        \toprule
        Dataset & Nodes & Edges & Classes & Feature Dimensions \\
        \midrule
        Cora     & 2708  & 5429  & 7       & 1433 \\
        PubMed   & 19717 & 44338 & 3       & 500  \\
        Citeseer & 3327  & 4732  & 6       & 3703 \\
        \bottomrule
    \end{tabular}
    \label{tab:dataset_statistics}
\end{table}

\subsection{Baselines}
We compared RW-NSGCN with several models, which can be broadly categorized into four classes. 
\textbf{Aggregation-based Models:} Aggregation-based models focus on updating node representations by aggregating features from neighboring nodes to capture both local and global graph structures. GCN~\cite{Kipf_Welling_2016,gan2022multigraph} uses spectral graph convolutions, GraphSAGE~\cite{Hamilton_Ying_Leskovec_2017, Chen2021Adversarial} employs sampling and aggregation for inductive learning, and GATv2~\cite{Brody_Alon_Yahav_2021, Shi2021Boosting-GNN} integrates an attention mechanism for dynamic feature weighting. PGCN~\cite{Ying_He_Chen_Eksombatchai_Hamilton_Leskovec_2018} extends this class by incorporating PageRank-based node centrality measures, highlighting influential nodes and personalizing parameter settings during the convolution process.
\textbf{Robust Representation and Sampling Techniques:}This class of models enhances graph representation learning through robust sampling strategies and sophisticated techniques to combat common challenges like over-smoothing. MCGCN~\cite{Yang_Ding_Zhou_Yang_Zhou_Tang_2020} employs Monte Carlo methods for effective negative sampling, while D2GCN~\cite{Duan_Xuan_Qiao_Lu_2022} and SDGCN~\cite{duan2023graph} focus on selecting diverse negative samples, improving robustness and maintaining distinct node representations. These approaches ensure more discriminative and robust embeddings by exploring the graph's latent structure thoroughly.
\textbf{Heterogeneous and Multi-Relational Graph Models:}Heterogeneous and multi-relational models are designed to handle complex graph structures with diverse node types and multiple relation types. R-GCN~\cite{Kim_Oh_2022} excels in managing multi-relational data through relation-specific weight matrices, making it adept for applications involving knowledge graphs and social networks. 
\textbf{Over-smoothing Mitigation and Depth Enhancement:}Models in this category focus on mitigating over-smoothing and enabling deeper architectures through innovative strategies. MAD~\cite{Chen_Lin_Li_Li_Zhou_Sun_2020} introduces a topological regularizer to maintain feature diversity, while DGN~\cite{Zhou_Huang_Li_Zha_Chen_Hu_2020} employs group normalization techniques to stabilize feature distributions, facilitating effective training of deeper networks. These models address critical challenges in deep graph neural networks, ensuring both depth and diversity in node representations.

\subsection{Experiment Settings}
In this experiment, we assessed the effectiveness of the model on the Cora and CiteSeer datasets and extracted subgraphs from the PubMed dataset to introduce artificial topological and weight perturbations for robustness testing. For all nodes, we implemented negative sampling with a maximum displacement of five. Candidate nodes were selected from second-order, third-order, and fourth-order neighbors.  Due to the varying structures, feature distributions, and category distributions of the different datasets, we negatively sample all nodes in the Cora and CiteSeer datasets, while for the PubMed dataset, we negatively sample nodes with node degrees between 3 and 6 (about 10\% of the overall number of nodes) in order to reduce the computational effort. Each dataset underwent 200 training iterations, and model parameters were selected from the best-performing epoch for testing. During training, we used the Adam optimizer with a learning rate of 0.01 and conducted ten tests per model on each dataset.

\subsection{Metrics}

\subsubsection{Accuracy}

Accuracy is used to assess the performance of a classification model, indicating the proportion of correctly classified instances within the dataset. The formula for calculating accuracy is:

\begin{equation}
    \text{Accuracy} = \frac{\text{Number of Correctly Classified Instances}}{\text{Total Number of Instances}}
\end{equation}

\subsubsection{Mean Average Distance}

The Mean Average Distance (MAD) calculates the average absolute difference between vectors and measures the overall variability through the cosine distance, reflecting the dispersion of the data.

MAD is defined as follows:

\begin{equation}
\text{MAD} = \frac{\sum_i D_i}{\sum_i \frac{1}{D_i}}, \quad D_i = \frac{\sum_j D_{ij}}{\sum_j \frac{1}{D_{ij}}},
\end{equation}
where $D_{ij} = 1 - \cos(x_i, x_j)$ denotes the cosine distance between nodes $i$ and $j$.A smaller MAD value can aid in clustering nodes of the same category more closely, but if too small, it may cause confusion between different categories. Conversely, a larger MAD value can help in distinguishing different categories but may cause nodes of the same category to be too dispersed. In short, lower MAD values indicate higher compactness, while higher MAD values indicate better separability.

\subsection{Experiment Results}

To validate the generalizability of the model, Table~\ref{tab:combined_results} presents the accuracy and Mean Average Distance (MAD) of various models on the Citeseer, Cora, and PubMed datasets. It is evident that our model significantly outperforms others in accuracy, achieving substantial improvements over all other models. This performance disparity primarily stems from differences in model structures. For example, GCN, MAD, and DGN employ fixed adjacency aggregation, while GraphSAGE's sampling strategy leads to insufficient information capture. Moreover, the random negative sampling strategies of D2GCN and SDGCN fail to capture crucial nodes, resulting in poor performance on complex graphs. On the PubMed dataset, these limitations are particularly evident, as models such as GCN and GraphSAGE struggle to effectively capture the intricate relationships in the data. In contrast, RW-NSGCN exhibits excellent performance, which may be attributed to its multilevel information aggregation and negative sampling mechanism that effectively captures global and local graph features and enhances the adaptability to noise and perturbation. 

In the Cora and PubMed datasets, our model exhibits low MAD values, indicating tighter clustering of intra-class nodes in the feature space. This contributes to greater similarity among community nodes and improves the model's robustness when handling noisy or uncertain data. Despite the low MAD values, our model does not suffer from over-smoothing. This is due to the integration of the Random Walk with Restart (RWR) algorithm and the Determinantal Point Process (DPP)-based GCN, which together maintain feature consistency while preventing overly similar representations among different nodes, thereby preserving the precision and discriminative power of classification decisions. In the Citeseer dataset, our model's MAD value is higher than that of most competing models. This suggests that the model generates more diverse feature representations on the Citeseer dataset, with greater differences between features. Nevertheless, our model still achieves the highest classification accuracy, demonstrating its ability to effectively leverage these feature differences to improve decision-making accuracy. The observed variation in MAD values likely stems from the complexity of the Citeseer dataset and the substantial heterogeneity among nodes, where greater feature diversity contributes to improved classification performance.This variation in MAD values reflects the structural strengths of our model. The model effectively maintains feature consistency to avoid over-smoothing while also improving feature diversity when necessary. This dynamic adjustment capability enables the model to achieve optimal classification performance across different scenarios.

\begin{table}[ht]
\centering
\caption{This table presents a comparative analysis of accuracy and Mean Average Distance (MAD) for various four-layer models on the Citeseer, Cora, and PubMed datasets. The evaluated models include GCN, GraphSAGE, GATv2, RGCN, MCGCN, PGCN, D2GCN, MAD, DGN, SDGCN, and RW-NSGCN. Performance metrics are reported as mean ± standard deviation.}
\label{tab:combined_results}
\begin{tabular}{lcccccc}
\toprule
 & \multicolumn{3}{c}{Accuracy} & \multicolumn{3}{c}{MAD} \\
\cmidrule(r){2-4} \cmidrule(r){5-7}
Model & Citeseer & Cora & PubMed & Citeseer & Cora & PubMed \\
\midrule
GCN & 51.41$\pm$5.94 & 61.7$\pm$5.89 & 73.87$\pm$5.05 & 66.22$\pm$9.47 & 69.98$\pm$9.37 & 82.58$\pm$11.19 \\
GraphSAGE & 61.25$\pm$4.60 & 73.1$\pm$4.50 & 75.01$\pm$2.16 & 77.75$\pm$6.52 & 69.13$\pm$6.06 & 85.35$\pm$8.02 \\
GATv2 & 56.57$\pm$7.30 & 69.7$\pm$4.40 & 73.39$\pm$2.22 & 76.72$\pm$6.35 & 73.88$\pm$5.11 & 80.7$\pm$11.09 \\
RGCN & 51.72$\pm$9.52 & 52.14$\pm$8.99 & 69.33$\pm$3.13 & 73.54$\pm$7.42 & 68.60$\pm$5.24 & 69.44$\pm$8.12 \\
MCGCN & 47.99$\pm$5.24 & 67.37$\pm$6.22 & 73.28$\pm$3.49 & 58.03$\pm$7.28 & 67.92$\pm$2.48 & 77.98$\pm$6.28 \\
PGCN & 56.52$\pm$9.31 & 58.68$\pm$9.77 & 69.83$\pm$3.19 & 74.87$\pm$7.51 & 66.71$\pm$5.85 & 76.42$\pm$4.87 \\
D2GCN & 61.30$\pm$3.57 & 71.75$\pm$4.04 & 76.45$\pm$0.77 & 72.82$\pm$4.80 & 72.35$\pm$8.21 & 87.81$\pm$5.04 \\
MAD & 56.15$\pm$8.69 & 67.49$\pm$8.62 & 71.15$\pm$5.30 & 79.91$\pm$3.50 & 75.46$\pm$5.54 & 80.86$\pm$10.12 \\
DGN & 63.76$\pm$3.97 & 71.63$\pm$3.09 & 77.18$\pm$2.37 & 87.70$\pm$3.46 & 84.17$\pm$7.50 & 87.71$\pm$6.29 \\
SDGCN & 65.38$\pm$3.26 & 74.21$\pm$1.41 & 75.26$\pm$1.33 & 83.53$\pm$1.18 & 81.34$\pm$3.65 & 90.41$\pm$3.54 \\
RW-NSGCN & \textbf{69.37$\pm$1.85} & \textbf{79.66$\pm$2.03} & \textbf{76.99$\pm$1.31} & \textcolor{gray}{80.21$\pm$1.01} & \textcolor{gray}{68.50$\pm$1.50} & \textcolor{gray}{77.93$\pm$1.04} \\
\bottomrule
\end{tabular}
\end{table}

\subsection{Graph Structure Perturbations}

In graph-structured networks, perturbations in topology and edge weights are inherent characteristics due to the dynamic and complex nature of these networks. We have designed two types of attacks—topological perturbation and weight perturbation—to amplify these natural disturbances, simulating more extreme scenarios to evaluate the model's robustness and stability under such conditions.

\begin{itemize}
    \item \textbf{Topological Perturbation:} This attack selectively removes critical edges, intensifying the naturally occurring topological variations within the graph network. It simulates the fragmentation and connectivity disruption that might occur under extreme circumstances, aiming to assess how well the model performs when faced with significant structural disturbances and to evaluate its capacity to handle complex changes in real-world networks.
    \item \textbf{Weight Perturbation:} This attack modifies edge weights or introduces random noise to amplify the existing fluctuations in edge weights within the graph. By simulating extreme weight disturbances, this approach assesses the model's robustness in handling changes in information flow paths and variations in node influence, ensuring the model's reliability in dynamic environments.
\end{itemize}

\subsection{Comparative Experiments with State-of-the-Art Models under Attack Conditions}

\begin{figure}
    \centering
    \includegraphics[width=1\linewidth]{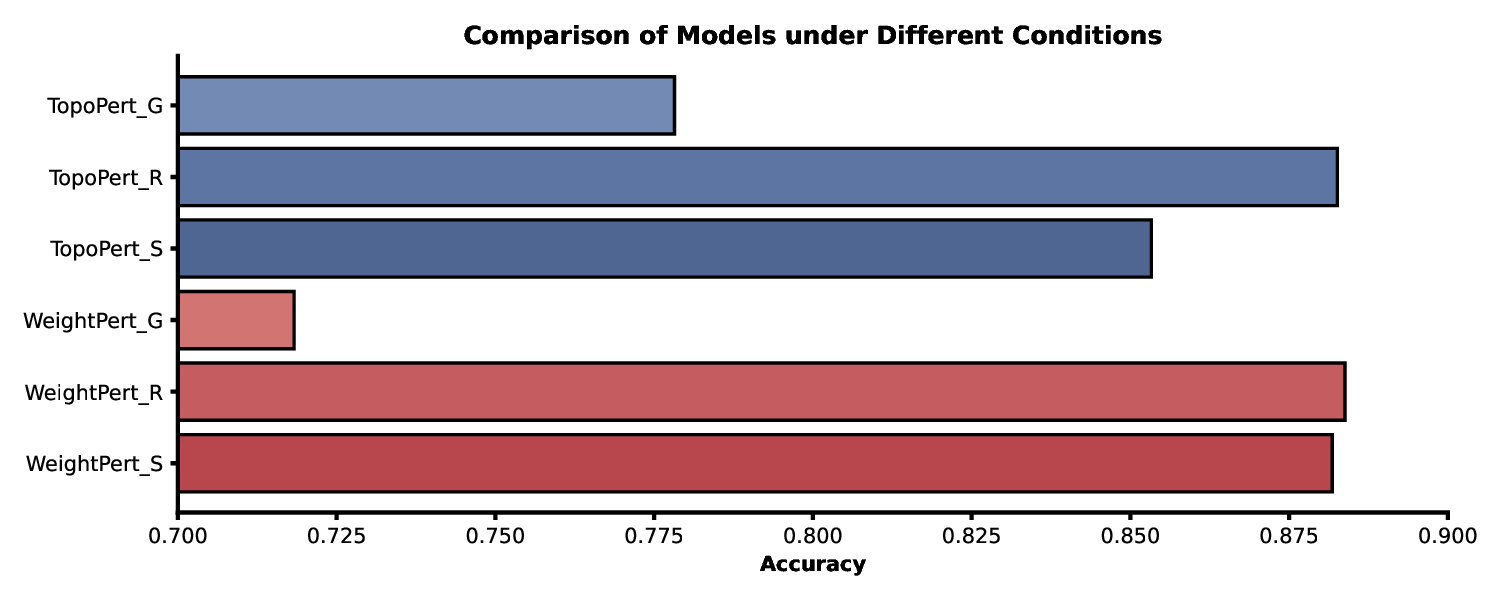}
    \caption{The bar chart illustrates the accuracy of three representative models in classifying network structures after two types of artificial perturbations: topological perturbation (CTBCA) and weight perturbation (TWPA). 'S' denotes the SDGCN model, the state-of-the-art model, 'R' represents the RW-NSGCN model developed in this study, and 'G' stands for the classic GCN model. The horizontal axis represents classification models, and the vertical axis represents accuracy. The height of the bars indicates the accuracy of each model under specific conditions.}
    \label{fig:attack}
\end{figure}

To evaluate the robustness of the RW-NSGCN model against adversarial attacks, we conducted comparative experiments with leading state-of-the-art models. The results are presented in Figure~\ref{fig:attack}.

The Graph Convolutional Network (GCN) applies convolution operations on graph structures to capture node relationships. However, when subjected to adversarial attacks, GCN struggles to accurately aggregate node features during convolution. Specifically, under the Topology Weight Perturbation Attack (TWPA), GCN achieves an accuracy of 0.7183, while under the Constant Topology and Bias Change Attack (CTBCA), its accuracy is 0.7782.

The Self-Diverse Graph Convolutional Network (SDGCN) combines GCN with Determinantal Point Processes (DPP) to reconstruct the network structure through negative sampling to manage perturbations. This method gives SDGCN an advantage in handling topological disturbances, reaching an accuracy of 0.8533. However, in weight perturbation scenarios, its accuracy decreases to 0.8818 due to incorrect edge weight information.

Building on SDGCN, the Random Walk-Node Sampling GCN (RW-NSGCN) improves the label propagation stage of GCN by integrating random walk and PageRank-based negative sampling techniques and optimizing sampling outcomes with DPP. This approach captures the global network structure by simulating random walks between nodes and using PageRank to assess node importance, while DPP ensures diversity and comprehensiveness in sampling. Consequently, it significantly improves the network's resilience. The RW-NSGCN model performs exceptionally well under both perturbation strategies, achieving accuracies of 0.8838 and 0.8826, respectively.

Overall, RW-NSGCN surpasses both SDGCN and GCN under various attack conditions, demonstrating superior robustness.

\subsection{Ablation Experiment}

To better understand the RW-NSGCN, we conducted ablation experiments to evaluate its effectiveness. As shown in Table~\ref{fig:abla}, the RW-NSGCN model achieved accuracies of 69.37 and 79.66 on the Citeseer and Cora datasets, respectively, surpassing models that use a single technique. Moreover, the RW-NSGCN model performed well in terms of Mean Average Distance (MAD), with MAD values of 80.21 for the Citeseer dataset and 68.50 for the Cora dataset. This indicates that the RW-NSGCN model has an advantage in reducing overfitting risks. The PageRank and random walk algorithms capture the global and local features of graph-structured data, respectively. By effectively integrating these techniques, the RW-NSGCN model not only improves node classification accuracy but also improves its ability to handle complex graph data.

\begin{table}[ht]
    \centering
    \caption{This table presents the results of an ablation study, comparing the performance of three models: PGR-GCN, RWR-GCN, and RW-NSGCN. The PGR-GCN model utilizes only the PageRank (PGR) algorithm, whereas the RWR-GCN model employs the Random Walk with Restart (RWR) algorithm exclusively. The RW-NSGCN model integrates both the PGR and RWR algorithms. The table contrasts the accuracy and mean absolute deviation (MAD) of these models on the Citeseer and Cora datasets.}
    \label{fig:abla}
    \begin{tabular}{lcccc}
        \toprule
        \textbf{Model} & \multicolumn{2}{c}{\textbf{Accuracy}} & \multicolumn{2}{c}{\textbf{MAD}} \\
        \cmidrule(lr){2-3} \cmidrule(lr){4-5}
         & \textbf{Citeseer} & \textbf{Cora} & \textbf{Citeseer} & \textbf{Cora} \\
        \midrule
        PGR-GCN  & 65.67 $\pm$ 2.60 & 78.45 $\pm$ 0.60 & 73.73 $\pm$ 0.21 & 79.31 $\pm$ 0.13 \\
        RWR-GCN  & 66.09 $\pm$ 1.42 & 77.69 $\pm$ 1.81 & 86.09 $\pm$ 0.15 & 65.51 $\pm$ 0.12 \\
        RW-NSGCN & 69.37 $\pm$ 1.85 & 79.66 $\pm$ 2.03 & 80.21 $\pm$ 1.01 & 68.50 $\pm$ 1.50 \\
        \bottomrule
    \end{tabular}
\end{table}

\begin{figure}[h]
\centering  
\includegraphics[width=1\linewidth]{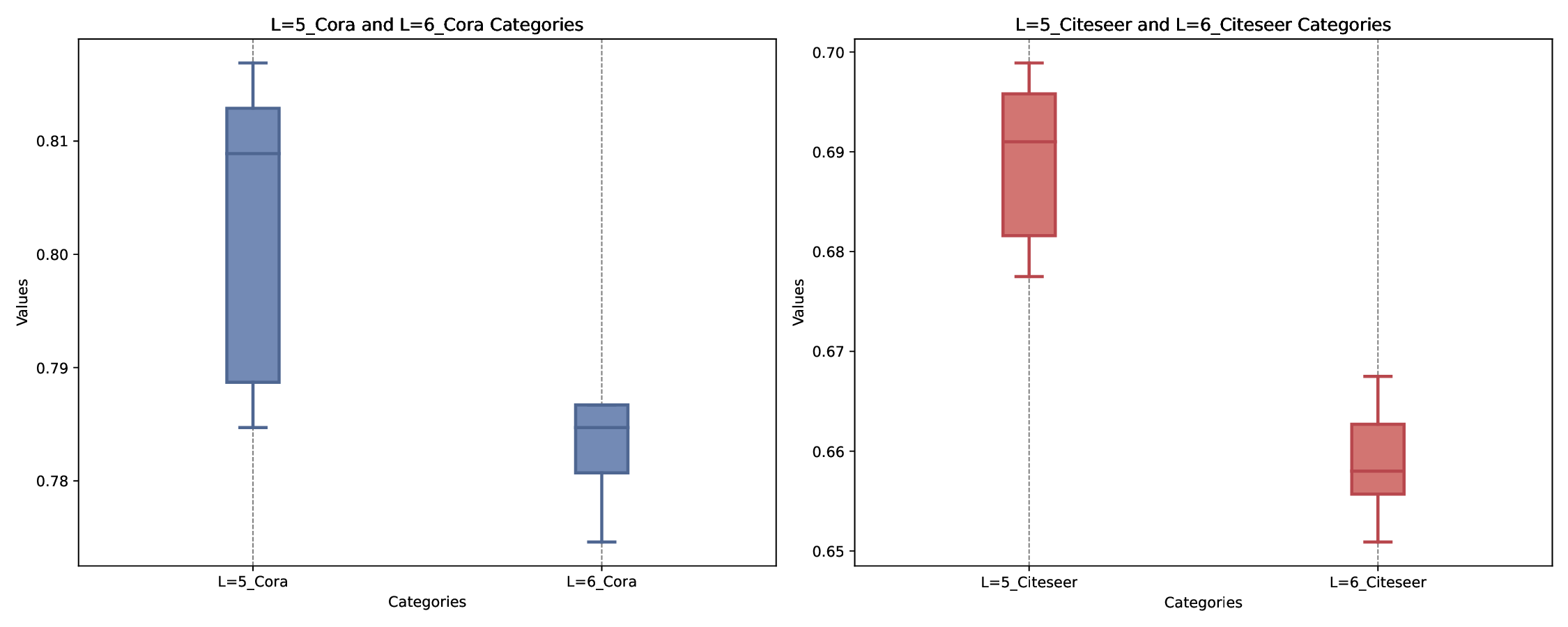}  
\caption{This figure illustrates the comparative accuracy of node selection using different $L$ values ($L=5$ and $L=6$) on the Cora and Citeseer datasets. Here, $L$ represents the maximum distance moved to select non-neighboring nodes. The left image shows that in the Cora dataset, the model with $L=5$ demonstrates significantly higher accuracy, despite some fluctuations. Meanwhile, the right image indicates that in the Citeseer dataset, the model with $L=5$ displays slightly higher accuracy than $L=6$ but with greater variability.}  
\label{fig:L}
\end{figure}

\subsection{Sensitivity Analysis of Maximum Non-Neighbor Distance}
According to Figure~\ref{fig:L}, the comparison of model accuracy under different parameter settings ($ L=5 $ and $ L=6 $) on the Cora and Citeseer datasets is shown. This analysis clearly demonstrates the superiority of the $ L=5 $ setting. Under this configuration, the model exhibits higher accuracy and more consistent performance across both datasets, indicated by a narrower interquartile range, highlighting the model's robustness. In contrast, when the parameter is set to $ L=6 $, there is lower accuracy and a wider distribution of data points, suggesting reduced stability and effectiveness at this parameter level. This comparison highlights the advantage of the $ L=5 $ setting in providing better and more reliable model performance.

\section{Conclusion}\label{sec6}
This paper introduces a novel approach that combines Restart Random Walk (RWR) and PageRank algorithms for negative sampling and employs a Graph Convolutional Network (GCN) based on Determinantal Point Processes (DPP) to address the topological vulnerabilities and weight instability present in Graph Neural Network (GNN) classification tasks. The model generates non-neighbor node sets across different path lengths based on the shortest path between nodes and uses RWR and PageRank to measure node importance. DPP ensures diversity among selected nodes, while GCN aggregates information, improving the model's adaptability and stability against topological vulnerabilities and weight perturbations in graph structures. Experimental evaluations demonstrate that RW-NSGCN excels in addressing topological and weight perturbations in graph structures, showing strong robustness and outperforming current state-of-the-art models.

\bibliography{sn-bibliography}

\end{document}